\documentclass{article}
\usepackage{ijcai19}

\usepackage{times}
\usepackage{soul}
\usepackage{url}
\usepackage[hidelinks]{hyperref}
\usepackage[utf8]{inputenc}
\usepackage[small]{caption}
\usepackage{graphicx}
\usepackage{amsmath}
\usepackage{booktabs}
\usepackage{algorithm}
\usepackage{algorithmic}
\title{RLTM: An Efficient Neural IR Framework for Long Documents}

\author{
Chen Zheng \and
Yu Sun \and
Shengxian Wan \and
Dianhai Yu
\affiliations
Baidu Inc., Beijing, China
\emails
\{zhengchen02, sunyu02, wanshengxian, yudianhai\}@baidu.com
}

\begin{document}
\maketitle

\begin{abstract}
Deep neural networks have achieved significant improvements in information retrieval (IR). However, most existing models are computational costly and can not efficiently scale to long documents. This paper proposes a novel End-to-End neural ranking framework called Reinforced Long Text Matching (RLTM) which matches a query with long documents efficiently and effectively. The core idea behind the framework can be analogous to the human judgment process which firstly locates the relevance parts quickly from the whole document and then matches these parts with the query carefully to obtain the final label. Firstly, we select relevant sentences from the long documents by a coarse and efficient matching model. Secondly, we generate a relevance score by a more sophisticated matching model based on the sentence selected. The whole model is trained jointly with reinforcement learning in a pairwise manner by maximizing the expected score gaps between positive and negative examples. Experimental results demonstrate that RLTM has greatly improved the efficiency and effectiveness of the state-of-the-art models.
\end{abstract}

\section{Introduction}
Ranking models based on deep neural networks have achieved significant improvements in information retrieval (IR).
Given a query and a list of documents, the goal of a ranking model is to generate a list of ranking scores, which can well present the relevance matching degree of the documents and the corresponding query. 
Compared to handcrafted features based learning to rank models where feature engineering is usually time-consuming, models based on deep learning can automatically learn features from raw texts of query and document.

Deep learning based ranking models mainly include two classes: representation based models, such as DSSM~\cite{huang2013learning} and CDSSM~\cite{shen2014learning} and interaction based models, such as DRMM~\cite{guo2016deep}, MV-LSTM~\cite{wan2016deep}, K-NRM~\cite{xiong2017end}, Conv-KNRM~\cite{dai2018convolutional}, MatchPyramid~\cite{pang2016text}, Match-SRNN~\cite{wan2016match}, PACRR~\cite{hui2017pacrr} etc. Representation models prefer to directly obtain the representations of query and document and generate the relevance score through the similarity between the representations, meanwhile, interaction models further consider the local interaction between terms of a query and a document. 
These models are mainly based on deep neural networks, such as Convolutional Neural Network~\cite{lecun1999object} and Recurrent Neural Networks~\cite{graves2013speech},  and also based on the interactions between query terms~\cite{fang2004formal} and documents terms.
Although these models achieved good performance, they are usually computationally expensive and are not efficient to scale to long documents.
This is mainly because the time complexity of these models is linear or quadratic respect to the query and document length, which may range from a few hundred to many thousands.

However, the vast majority of these studies have overlooked a problem: Not every sentence in a source long document has the same importance for relevance matching. Consider the human judgment process~\cite{wu2007retrospective}, given a query and a document, the human annotator firstly skim the whole document quickly to locate the most relevant part of the document, and then the human annotator matches the query with the selected parts carefully to decide the relevance label.
Based on this observation, we can reduce the computational burden through selecting relevant sentences to prune down long document.
DeepRank~\cite{pang2017deeprank} also adopts this idea to model the matching for long documents.
However, DeepRank selects relevant parts from the documents in a trivial way, which is based on hand-crafted rules.
The trivial handcrafted rules may lead to extracting wrong parts in this step and the errors will make a big limit to the final performance. Same as PACRR~\cite{hui2017pacrr}, which described two strategies: first$k$ and $k$window. In fact, PACRR-first$k$ simply keeps the first k terms in the document without knowing all the document information, and PACRR-$k$window only chooses top $k$ terms with the simple position's similarity measure. If we extract the wrong parts from a document in the first step, how could we expect any model can obtain the correct relevance degree in the next steps?

In this paper, we propose a novel Reinforced End-to-End neural IR framework  for  Long Text Matching (RLTM).
The framework combines with two models, a sentence selection model and a sentence matching model.
The sentence selection model selects relevant sentences in the long documents by an efficient sentence selection model.
Then, the sentence matching model generates a relevance score by a more sophisticated matching model based on the sentence selected.
In order to avoid the error accumulation problem, the two models are trained jointly with reinforcement learning.
By joint training, the two models can learn to cooperate with each other during the training process.
Specifically, the sentence selection model will learn to select the sentences with the best discriminability for the sentence matching model to discriminate between the positive and negative documents.
We can consider our framework as a reinforcement learning agent, where the state is a query and a document, and the actions are which sentences to select from the document, and the reward is obtained by discriminating the positive and negative documents.
The framework is trained pairwise and the training object of reinforcement learning is to maximize the expected relevance score gap between positive and negative documents.
To the best of our knowledge, the RLTM framework is the first time to train the whole model jointly with reinforcement learning for the relevance matching task.

We conduct extensive experiments based on two datasets, a human-labeled dataset and a click-through dataset, and compare our framework with state-of-the-art IR models.
Experimental results show that the RLTM framework not only achieves higher accuracy but also accomplish lower computational cost compared to the baselines. 

\section{Related Work}
\subsection{Learning to Rank}
Over the years, with the rapid development of machine learning and deep learning, Learning to Rank(LTR) area has also made tremendous progress. The essence of LTR can be divided into three categories: Point-wise, Pair-wise and List-wise methods. These three different methods correspond to three different input and output, and three different loss functions. Point-wise methods take document vectors as input and the ranking score as output, and logistic regression~\cite{gey1994inferring} is the most representative of this method. Pair-wise methods, like RankSVM~\cite{joachims2002optimizing}, take a positive document and negative document pair as input and generate the ranking score pair as output, which loss function usually take hinge loss as loss function. List-wise methods take a list of documents which match the same query as input and generate a list of ranking scores, such as ListNet~\cite{cao2007learning} and AdaRank~\cite{xu2007adarank}. In this paper, we consider the Pair-wise method for our experiment.

\begin{figure*}[ht!]
\centering
\includegraphics[width=0.8\textwidth,height=0.3\textheight]{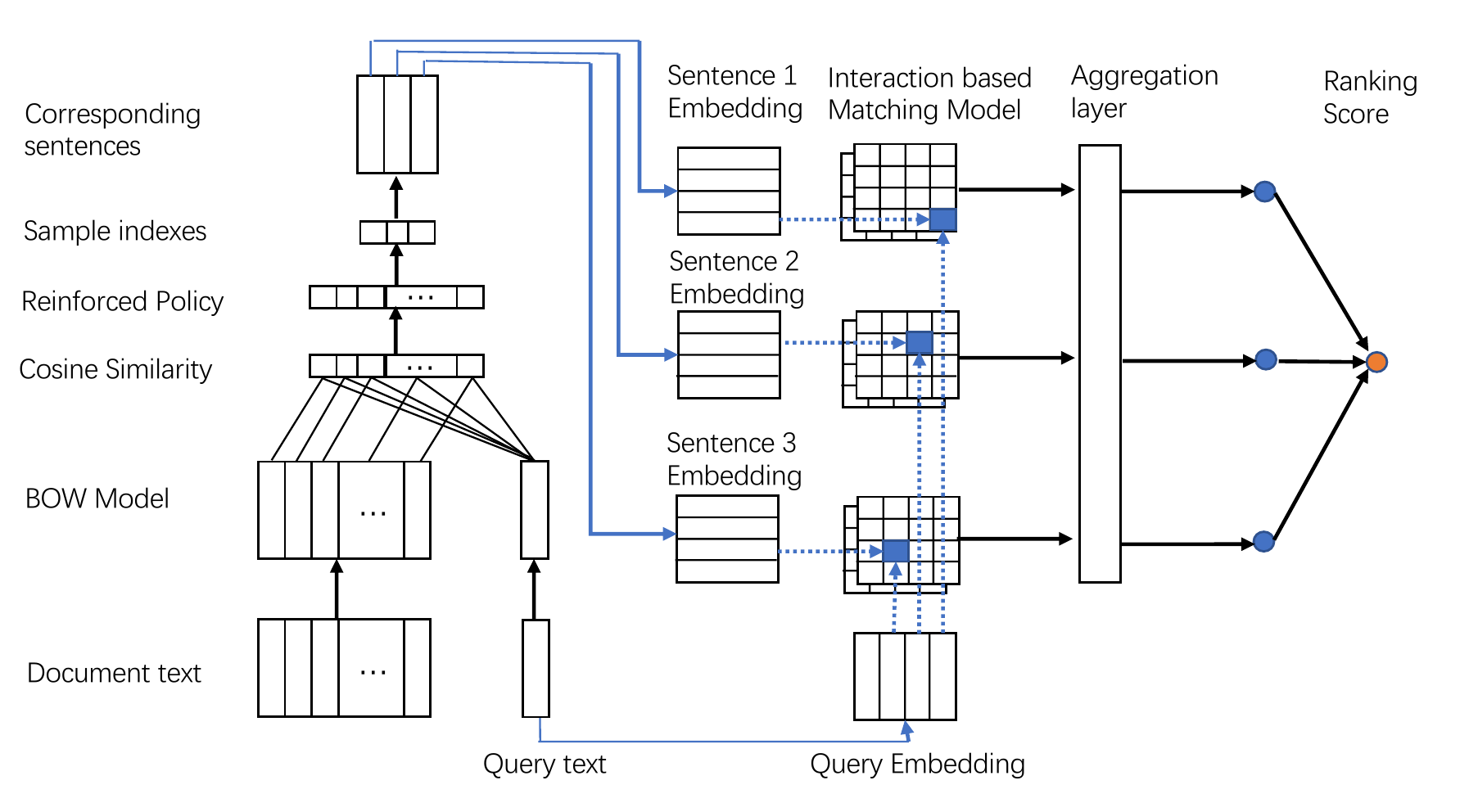}
\caption{The framework of Reinforced Long Text Matching (RLTM). We choose 3 sentences from sentence selection model as the example for our framework figure. }
\end{figure*}

\subsection{Neural IR Models}
In recent years, with the development of IR research, neural IR models can be divided into two classes: representation based IR models and interaction based IR models~\cite{guo2016semantic}. For the representation based IR models, especially like DSSM~\cite{huang2013learning} and CDSSM~\cite{shen2014learning}, these models usually learn that how to generate a good representation for query and document separately, and then compute a simple similarity score based on dense representations, such as cosine similarity methods. However, these models ignore the importance of exact matching signal and the significant of relevance matching. Furthermore, these models only consider that how to build up directly representations between the document and query. For the interaction based IR models, such as DRMM~\cite{guo2016deep}, MV-LSTM~\cite{wan2016deep}, K-NRM~\cite{xiong2017end}, MatchPyramid~\cite{pang2016text}, Match-SRNN~\cite{wan2016match}, PACRR~\cite{hui2017pacrr}, these models are more inclined to further consider the local interaction between terms of a query and a document. In addition to these two classes of models, a hybrid version of the model, named DUET model~\cite{mitra2017learning}, combines the representation based model and interaction based model.

However, regardless of the first class of representation based model or the second class of interaction based model, they all overlook a problem: Not every sentence in a source long document has the same importance for relevance matching. The work of DeepRank~\cite{pang2017deeprank} proposed a deep learning model to handle long document, by finding the relevant location, measuring the local relevance, and determining the relevance score. However, DeepRank selects relevant parts from the documents in a trivial way, which is based on hand-crafted rules. This trivial handcrafted rules may extract wrong parts from the long document and even produce the negative influence on the final performance.

\section{Reinforced Long Text Matching (RLTM) Framework}
In this section, we first describe the high-level overview of our End-to-End framework for long text matching (RLTM). 
Furthermore, we separately introduce the implementation of the two components of RLTM.

\subsection{High-Level Overview of RLTM}
Given a query $q$ and a document $d$, the object of RLTM is to output a relevance score for them.
A document can be considered as a collection of semantic units (e.g. paragraphs or sentences). 
Without loss of generality, in this paper we take sentences as the semantic units and $d = \{u_1, u_2, ..., u_T\}$, where $T$ is the sentence count of the document.
In general,  RLTM consists of two parts: a sentence selection model  and a fine grained sentence matching model, as shown in Figure 1. 
The sentence selection model compares each sentence in the document with the query and outputs a policy of how to select the sentences, i.e., a probabilistic distribution $\pi(u|q, d)$ over the sentences,
\begin{equation*}
\begin{aligned}
& P = \mbox{softmax}(\Phi(q, u_1),...,\Phi(q, u_T)) , \\
& \pi(u = u_t | q, d) = P_t,
\end{aligned}
\end{equation*}
where $\Phi(q, u)$ is a matching model which outputs the relevance score between $q$ and $u$ before normalization.
According to the descending order of the probabilistics, we can select the most $K$ important sentences,  notated as $U' = \{u'_1, ..., u'_K\}$, for the following comparison. 
Then, the sentence matching model takes the query and the sentences selected as inputs, and outputs a overall relevance score for the whole documents, 
\begin{equation*}
s = F(q, U') = \Lambda( \Psi(q,u'_1) ,..., \Psi(q,u'_K)),
\end{equation*}
where $\Psi(q, u)$ is also a matching model which outputs the relevant representation between $q$ and $u$, $\Lambda$ is a aggregation function to aggregate the relevance representations of all the selected sentences and outputs the final relevance score between the query and the document (i.e.~$s$).
In practice, $\Psi$ is more sophisticated and computation expensive than $\Phi$ and thus our framework can reduce the computation significantly without loss of performance.
In general, this framework is not restricted to specific matching models, $\Phi$ and $\Psi$, we will describe the models  implemented in this paper in next sections.

\subsection{Sentence Selection Model}
In this section, we describe the implementation of the function $\Phi$, which matches each sentence with the query in order to select the most important sentences.
Since all the sentences in the document are considered, this model should be effective and efficient.
In this paper, we adopt the BoW model which represents a text by averaging the embeddings of all the terms in the text.
There are several advantages to the BoW model.
Firstly, comparing to sophisticated matching models, such as CNN and RNN based matching models, BoW is extremely efficient.
Secondly, BoW is an embedding based model which conducts matching at the semantic level.
Lastly, BoW is also very flexible with a huge number of embedding parameters that can be learned End-to-End.

Specifically, to obtain the representation for each text, $u = \{x_1, x_2, ..., x_L\}$, we firstly use BoW model which averages all the term embeddings $\{e_1, e_2, ..., e_L\} $ and then the result is transformed to a new semantic space by a nonlinear full connected neural layer.

Afterward, cosine similarity is adopted to obtain the relevance score between the query $q$ and the sentence $u$. 
Finally, we use softmax operation to normalize the relevance score overall sentences in a document.
\begin{equation*}
\begin{aligned}
&h_q = \mbox{tanh}(W_q(\mbox{BoW}(q)) + b_q), \\
&h_{u} = \mbox{tanh}(W_u(\mbox{BoW}(u)) + b_u),\\
&c = \mbox{cos}(h_q, h_{u}) = \frac{h_q^T h_{u}}{||h_q|| \cdot ||h_{u}||},\\
&\pi(u=u_t | q, d) = \frac{e^{c_t}}{\sum_{k}{e^{c_k}}},
\end{aligned}
\end{equation*}
where $W_q, W_u, b_q, b_u$ are the parameters. Since the sentence selection model is not differentiable, with respect to the supervised loss commonly used, we use reinforcement learning to train this model, as shown in the next section.

\subsection{Fine-grained Sentence Matching Model}
In this section, we describe the implementation of the function $\Psi$ and $\Lambda$.
Since that the object in this step is to compare the selected sentences with the query in fine grained, for $\Psi$ we adopt two existing state-of-the-art matching models: MatchPyramid and K-NRM.
The two models are interaction based matching models which model the interaction between two texts at term level.
The document level relevance is aggregated from the fine grained term level matching signals.

MatchPyramid firstly builds up a matching matrix through term level matching signals between a query and a document.
Afterward, a convolutional neural network is adopted to extract different levels of matching patterns in the matching matrix.
Finally, the relevance score is generated through a full connection neural layer. 
In our experiment, we use cosine similarity as the interaction function~\cite{pang2016text}.

K-NRM is a kernel based neural ranking model~\cite{xiong2017end}. 
Like MatchPyramid, a translation layer is adopted to build a term level matching matrix.
Then, soft-TF counts are generated as ranking features by a kernel pooling layer. 
Finally, the relevance score is also generated through a full connection neural layer.

For $\Lambda$, while some complex aggregation functions such as neural networks can be adopted, for simplicity, we sum all the relevance scores generated by $\Psi$ in this paper. 

\section{Pairwise End-to-End Learning with Policy Gradient}
We describe the learning method for RLTM in this section.
Since that the sentence selection model is non-differentiable with respect to the supervised loss used in traditional supervised learning, we used reinforced learning to learn the parameters.
The action is which sentences to select by the sentence selection model for the sentence matching model and the policy function is $\pi(u | q, d)$.
Since the model is trained pairwise and the object is to discriminate between the positive and the negative documents, we define the reward as the relevance score gap between the positive and negative documents.
Given a  triple of query, a positive document and a negative document (i.e.~$(q, d^+, d^-)$), the reward function is defined as
$$R(U^+, U^-) = s^+ - s^-,$$
where $R(U^+, U^-)$ is the reward function, $U^+=\{u'^+_1,...,u'^+_K\},U^-=\{u'^-_1,...,u'^-_K\}$ are the sampled sentences, $s^+ , s^-$ are ranking scores.
Our Training object is to maximize the expectation of reward,
$$ J(\theta) = \mathbf{E}_{{U^+}\sim \pi({u}|q,d^+,\theta), {U^-}\sim \pi({u}|q,d^-,\theta)}R(U^+, U^-). $$

The whole training process is showed in Algorithm 1.

\begin{algorithm}
  \label{algorithm:alg1}
    \caption{Reinforced Long Text Matching (RLTM)}
  \begin{algorithmic}[1]
    \STATE \textbf{Input: }  dataset of ${(q, d^+, d^-)}$
    \STATE \textbf{Output:} $\theta$ 
    \STATE \textbf{Initialize:}
     $\theta \gets$ pre-trained $\theta$ by supervised learning shown in Sec 5.2.
    \FOR{ each ${(q, d^+, d^-)}$ in dataset}
      \STATE For $({q,d^+})$, randomly sample $K$ sentences from ${d^+}$ for training, $U^+ = \{u'^+_1, ..., u'^+_K\}\sim \pi (u | {q}, {d^+})$. 
      \STATE For $({q,d^-})$, randomly sample $K$ sentences from ${d^-}$ for training, $U^- = \{u'^-_1, ..., u'^-_K\} \sim \pi (u | {q}, {d^-})$. 
      \STATE Generate the ranking score $s^+, s^-$ by fine-grained matching model $F$
      \STATE Get reward $r=R(U^+, U^-)$ according to $s^+ - s^-$
      \STATE Updating the sentence selection model through policy gradient $r\sum_{k=1}^{K}[\frac{\partial}{\partial \theta}\log\pi(u'^+_k | q, d^+)+\frac{\partial}{\partial \theta}\log\pi(u'^-_k | q, d^-)]$
      \STATE Updating the Fine-grained matching model through gradient $\frac{\partial}{\partial \theta} {R(U^+, U^-)} $
  \ENDFOR
  \end{algorithmic}
\end{algorithm}

\section{Experiments}
In this section, we conduct a series of experiments to show the performance improvement against baselines. 

\begin{table}[!htbp]
\scriptsize
\centering
\begin{tabular}{c|c|c}
\hline \bf Statistics & \bf Human-Label & \bf Click-Through \\ \hline
Queries for train & 81922 & 37381 \\
Queries for validation & 6228 & 2951 \\
Queries for test & 7312 & 3426 \\
Avg Docs Per Query & 13 & 14.5 \\
Avg Doc Length & 2644.2 & 2724.0 \\
label & Human-annotated & DCTR \\
\hline
\end{tabular}
\caption{\label{font-table} Data statistics. }
\end{table}

\subsection{Experimental Setup}
\paragraph{Datasets.} We conduct our experiments on two large-scale datasets, both of them are from one Chinese search engine.
The first dataset, named Human-Label dataset, is a human-annotated dataset.
Each query-document pair is labeled with five level labels.
The second dataset, named Click-Through data, is sampled from the click-through search log. 
The search log records all searching, browsing and clicking behaviors of hundreds of millions anonymous users.
In the Click-Through dataset, we use DCTR click model to generate relevance labels through user clicks~\cite{chuklin2015click}. The DCTR labels have 5-grade labels from 0 to 4.

Because all of the documents are retrieved by the complex and complicate back-end algorithms by one Chinese search engine, these two datasets have very high quality. 
Each document is a Web page and we extract the title and body texts from the Web page for matching.
The title is an important representative of the document semantics and it is usually quite short.
In all of our experiments, we always select the title as the first sentence and the sentence selection model will select other sentences from the body texts.
The statistics of the datasets are listed in Table 1.

\paragraph{Implementation Details.} We implemented all the models using TensorFlow. Due to the model complexity constraint, we chose the most 1 million as the vocabulary size by word frequent of training corpus.  Meanwhile, we set word embedding dimension to 128 according to past experimental experience. We used stochastic gradient descent method, Adam\cite{kingma2014adam}, as our optimizer for the training. We set the batch size to 32 and selected the learning rate from [1e-1, 1e-2, 1e-3, 1e-4, 1e-5]. 

For the reinforced sentence selection model, the fully connected hidden size is 128, and we choose the number of sentences as $({1,3,5})$. For the fine-grained matching model, because we separately employ three different interaction based models, we will introduce their hypermeters individually. For MatchPyramid, we set the query window size to 2, sentence and sentence window size to 4. And the kernel size is 128. For K-NRM, we set the number of bins to 11. Those are the same settings using the hyper-parameters described in their respective papers.

\subsection{Evaluation Measure}
For our experiments, we use the common assessment methods in information retrieval and learning to rank field, mean average precision (MAP) and normalized discounted cumulative gain (NDCG), to evaluate our model and baseline models. We computed NDCG@1, NDCG@3, NDCG@5, and NDCG@10 separately. The formula of NDCG are as follows:
$$ NDCG(q,k) = \frac{1}{|q|}\sum_{j=1}^{|q|}Z_{j,k}\sum_{m=1}^k\frac{2^{R(j,m)}-1}{log(1+m)} $$
For the NDCG formula above, $Z_{j,k}$ means normalization factor, $R(j,m)$ means the score of relevance between the ith document's ranking result and the jth query. The value of k is {1, 3, 5, 10} in our experiments. 

\subsection{Baselines}
Our baselines include two classes of models: feature-based ranking models and neural ranking models. 

\paragraph{Feature-based Ranking Models.} This part of baselines include BM25, which is a popular unsupervised retrieval baseline for IR, and RankSVM, which is a strong feature-based and pairwise learning-to-rank baseline. Both BM25 and RankSVM use long document title and body.

\begin{table*}[ht!]
\scriptsize
\begin{center}
  \begin{tabular}{l|lllll|lllll}
    \hline
    Models &
      \multicolumn{5}{c}{Click-Through} &
      \multicolumn{5}{c}{Human-Label} \\ 
    & \bf NDCG@1 & \bf NDCG@3 & \bf NDCG@5 & \bf NDCG@10 & \bf MAP
	& \bf NDCG@1 & \bf NDCG@3 & \bf NDCG@5 & \bf NDCG@10 & \bf MAP \\
    \hline
BM25  & 0.235 & 0.343 & 0.400 & 0.478 & 0.413 &  0.300 & 0.473 & 0.484 & 0.507 & 0.443 \\
RankSVM & 0.239 & 0.350 & 0.406 & 0.481 & 0.420 &  0.302 & 0.477 & 0.488 & 0.509 & 0.454 \\
\hline
DSSM & 0.226 & 0.331 & 0.389 & 0.462 & 0.387 & 0.298 & 0.466 & 0.479 & 0.502 & 0.411 \\
CDSSM & 0.232 & 0.342 & 0.401 & 0.475 & 0.382 & 0.305 & 0.471 & 0.486 & 0.509 & 0.413 \\
\hline
MatchPyramid & 0.280 & 0.398 & 0.455 & 0.495 & 0.412  & 0.376 & 0.491 & 0.510 & 0.522 & 0.458 \\
K-NRM & 0.283 & 0.404 & 0.460 & 0.503 & 0.426 & 0.389 & 0.498 & 0.524 & 0.533 & 0.456  \\
\hline
DeepRank & 0.302 & 0.424 & 0.479 & 0.522 & 0.455 & 0.418 & 0.531 & 0.555 & 0.566 & 0.481 \\
\hline
RLTM-MP & 0.316 & 0.439 & 0.498 & 0.536 & 0.468 &  0.445 & 0.554 & 0.575 & 0.589 & 0.490 \\
RLTM-KNRM & 0.320 & 0.441 & 0.501 & 0.544 & 0.470 & 0.462 & 0.577 & 0.593 & 0.606 & 0.487 \\
\hline
\end{tabular}
\end{center}
\caption{Performance of the RLTM framework with all the baselines. The table shows the NDCG Result and MAP result for two test datasets and denoted significant improvement over all baseline models.}
\end{table*}

\paragraph{Neural Ranking Models.} This part of baselines includes DSSM, CDSSM, MatchPyramid, K-NRM and DeepRank. 
DSSM builds the representations of query and document by multilayer fully connected neural network, and finally generates the matching score by cosine similarity. CDSSM is the convolutional version of DSSM.

MatchPyramid and K-NRM directly take the full long documents as inputs and produce the relevance score.
Since the computation costs of MatchPyramid and K-NRM are linear to the document length, they are not efficient compared with RLTM.
DeepRank firstly selects relevance parts from the documents in a trivial way, which is based on handcrafting rules and only considers term exact match signal, and then adopts MatchPyramid and Match-SRNN to produce the relevance score. 




\subsection{Results and Analysis}
In this section, we will show the speed gain for our framework and the performance of our framework and a series of baselines on the two datasets. Furthermore, we will offer the detailed analysis of our framework.

\begin{figure}[!htbp]
\centering
\includegraphics[width=0.5\textwidth,height=0.25\textheight]{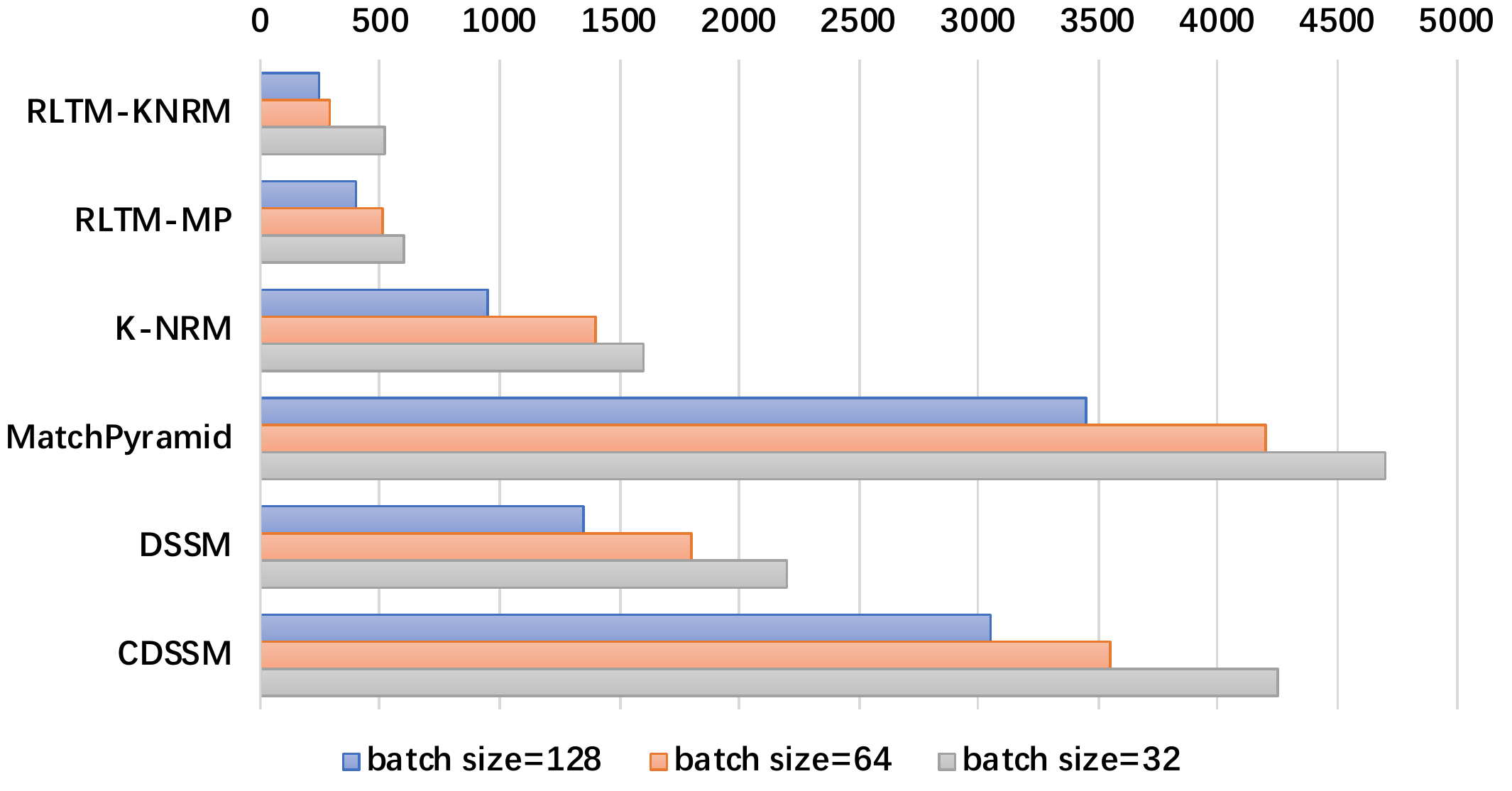}
\caption{Speed Gain: The test running time comparisons between our RLTM framework, MatchPyramid, K-NRM, DSSM and CDSSM. }
\end{figure}

\paragraph{Speed Gain for RLTM.} Since our RLTM framework chooses several sentences by the reinforced sentence selection model instead to handle with the whole long document, our framework has a significant acceleration effect compared with baselines in theory. 
we measure the inference time on human-label test data to give a detailed comparison.
We uniformly separately set batch size to 32, 64 and 128 for the inference time comparison between RLTM and other matching models which cope with the whole long document. Figure 2 shows us the speed gain. 
Experimental results show that RLTM-MP has 9.2 times faster acceleration than MatchPyramid, and RLTM-KNRM has 4.3 times of speed up compared to K-NRM. 

As the number of words in the content increases, the text matching matrix for MatchPyramid or CDSSM also becomes larger, resulting in a sudden increase in the computation time of the CNN layer's operation. The reason for the inefficient running time of DSSM is the Word Hashing and multi-layer neural network. Because DeepRank selects relevant parts from the documents in term exact matching signal, which is based on handcrafted rules, we do not show the running time of DeepRank in the figure2. 

This result demonstrates that our RLTM framework can better cope with ranking tasks with long documents, and is more suitable for applications in the real scenario, especially in industry.


\paragraph{Performance Results.} All experimental results are listed in Table 2.
From the results, we can get a series of observations and draw several conclusions.

Comparing feature-based models with representation based models, we can see that RankSVM slightly outperforms DSSM and CDSSM on both datasets. 
This demonstrates that although representation based models are good at matching at the semantic level, exact term matching signals are quite effective for relevance matching tasks.
Interaction based models, such as MatchPyramid and K-NRM, perform significantly better than representation based models and feature-based models.
This demonstrates that term level interactions are important signals for relevance matching.
End-to-End neural models are more powerful and flexible to learn the matching relationship of term pairs than manual features.
These conclusions are consistent with previous works~\cite{pang2017deeprank,xiong2017end}. We describe more details in Further Analysis for RLTM model.

Finally, our framework achieves the best performance over all baselines in both NDCG and MAP. The improvement of RLTM-KNRM against K-NRM is about $ 7.3\% $ on NDCG@1. More importantly, RLTM-KNRM is able to outperform DeepRank by more than $ 4.4\% $ on NDCG@1 and $ 4.0\% $ on NDCG@10.
In DeepRank, handcrafted rules may lead to extract wrong relevant parts and the errors will make a limit to the final performance. 
Comparing with DeepRank, the sentence selection model of RLTM is a more powerful and flexible embedding based model, which can learn the term matching relationship End-to-End.
Comparing with other baseline models, the sentence selection model in RLTM is trained by policy gradient, which directly optimizes the final reward.
Through joint training, the sentence selection model and sentence matching model can learn to cooperate with each other during the training process, and it can effectively solve the problems of other baseline models.

\paragraph{RL Strategy for RLTM Model.} In order to demonstrate the effectiveness of the reinforcement learning strategy adopted in our framework, we also introduce two other baselines, notated as pipeline models (Pipeline-MP and Pipeline-KNRM).
Pipeline models have the same inference process as RLTM framework.
Take Pipeline-MP as an example, Pipeline-MP also selects sentences by BoW model first and then uses MatchPyramid to match the sentences selected.
The difference lies in that pipeline models are trained in a pipeline way instead of jointly trained by reinforcement learning.
Firstly, the sentence selection model is trained the same as the pre-training step of RLTM.
Lastly, given the sentence selected by the trained sentence selection model, the sentence matching model is trained by traditional pairwise supervised learning. Table 3 shows us that RLTM models obtain the better performance for both NDCG and MAP compared with pipeline model. The improvement of RLTM-KNRM against Pipeline-KNRM is about $ 5.0\% $ on NDCG@1 in the human-label test dataset.

Comparing the pipeline models with original interaction based models(MatchPyramid and K-NRM) in table 2, although pipeline models conduct term level matching only for a subset term of the whole document, they perform better than interaction based models. This demonstrates that more does not mean better.
In most cases, the relevance degree can be determined by only considering the most important parts of the whole document.
If we take the long tail terms of the document into consideration, although they bring some benefits, they also bring many noisy irrelevant parts which will hurt the model during the training process. 
DeepRank is also a pipeline model and performs slightly better than other pipeline models. 
The reason is that DeepRank selects relevant parts from the documents based on term exact matching signals, which is trivial but fairly stable.
Other pipelines models train the sentence selection model with an indirect objective function which leads to a worse performance than DeepRank.

\begin{table}[ht!]
\tiny
\centering
\begin{tabular}{c|c|cccc}
\hline \bf Data & \bf Learning & \bf NDCG@1 & \bf NDCG@5 & \bf NDCG@10 & \bf MAP \\ \hline
Clickthrough & Pipeline-MP & 0.295 & 0.468 & 0.513 & 0.445 \\
& Pipeline-KNRM & 0.299 & 0.475 & 0.520 & 0.442 \\
& DeepRank & 0.302 & 0.479 & 0.522 & 0.455 \\
& RLTM-MP & 0.316 & 0.498 & 0.536 & 0.468 \\
& RLTM-KNRM & 0.320 & 0.501 & 0.544 & 0.470 \\
\hline
Human-Label & Pipeline-MP  & 0.400 & 0.538 & 0.551 & 0.475 \\
& Pipeline-KNRM & 0.412 & 0.551 & 0.560 & 0.472 \\
& DeepRank &0.418 & 0.555 & 0.566 & 0.481 \\
& RLTM-MP & 0.445 & 0.575 & 0.589 & 0.490 \\
& RLTM-KNRM & 0.462 & 0.593 & 0.606 & 0.487 \\
\hline
\end{tabular}
\caption{\label{font-table} Performance Comparison between RLTM-model, Pipeline-model and DeepRank.}
\end{table}

\paragraph{The Analysis of Reinforced Sentence Selection Model.} Table 4 shows us the performance of RLTM between choosing the different number of sentences in the reinforced sentence selection model. 
We take MatchPyramid as the fine-grained matching model to compare the performances. 
In this experiment, we individually sampled 1, 5, 7 sentences besides the title by the reinforced sentence selection model. 
Both of results from Human-Label and Click-Through datasets show us that choosing multiple sentences contribute to the fine-grained matching model. 
RLTM-MP-7 (choosing 7 sentences) gets the better performance for RLTM. 
However, RLTM-MP-7 will have quite limited improvement compared with RLTM-MP-5. 
Although the more selected sentences will help the local relevance judgment, the more probability of irrelevances will be happened and will influence the matching model.

\begin{table}[ht!]
\scriptsize
\centering
\begin{tabular}{c|c|ccc}
\hline \bf Data & \bf Learning & \bf NDCG@1 & \bf NDCG@10 & \bf MAP \\ \hline
Human-Label & RLTM-MP-1 & 0.430 & 0.572 & 0.475 \\
& RLTM-MP-5 & 0.441 & 0.587 & 0.487 \\
& RLTM-MP-7 & 0.445 & 0.589 & 0.490 \\
\hline
\end{tabular}
\caption{\label{font-table} The performance of RLTM model, which sentence selection model separately choose 1, 5, 7 sentences, and Fine-grained matching model choose MatchPyramid model. }
\end{table}

\section{Conclusion}
This paper presents a novel efficient neural IR framework, named RLTM, to cope with the long document relevance matching problem in information retrieval.
RLTM combines a sentence selection model and a fine-grained sentence matching model. 
We creatively use reinforcement learning to train our RLTM framework. 
To the best of our knowledge, it is the first time that using reinforcement learning for relevance matching problem. 
We conduct extensive experiments on both a human-labeled dataset and a click-through dataset.
Experimental results show that RLTM is effective and efficient.
RLTM not only outperforms the state-of-the-art baselines under relevance evaluation metrics, such as NDCG, but also is much faster than these models.

\bibliographystyle{named}
\bibliography{ijcai19}

\end{document}